\documentclass{article}

\usepackage{PRIMEarxiv}

\usepackage[utf8]{inputenc} % allow utf-8 input
\usepackage[T1]{fontenc}    % use 8-bit T1 fonts
\usepackage{hyperref}       % hyperlinks
\usepackage{url}            % simple URL typesetting
\usepackage{booktabs}       % professional-quality tables
\usepackage{amsfonts}       % blackboard math symbols
\usepackage{nicefrac}       % compact symbols for 1/2, etc.
\usepackage{microtype}      % microtypography
\usepackage{lipsum}
\usepackage{fancyhdr}       % header
\usepackage{graphicx}       % graphics
\graphicspath{{media/}}     % organize your images and other figures under media/ folder

\usepackage{algorithm}
\usepackage{algorithmic}
\usepackage{amsmath,amssymb}
\usepackage{multirow}
\usepackage{bbding}
\usepackage{subcaption}

\title{SAGE: Source-Anchored Guidance via Frequency Equalization for Hierarchical RGB-T Alignment and Fusion}

\author{
  Timing Li, Yiming Sun, Boan Tao, Xiyuan Gao, Haifang Cao,  Pengfei Zhu\\
}

\begin{document}
\maketitle

\begin{abstract}
Spatial misregistration and cross-modal discrepancies often cause ghosting, structural blurring, and content imbalance in RGB-T fusion. Existing methods typically decouple appearance adaptation, geometric alignment, and information fusion, limiting dependency propagation across stages. We propose Source-Anchored Guidance via Frequency Equalization for Hierarchical RGB-T Alignment and Fusion (SAGE), a unified framework integrating frequency equalization, hierarchical alignment, and subband fusion. SAGE employs invertible joint encoding and source-specific low-frequency modulation to derive structural and gain guidance while preserving source information. Hierarchical frequency collaborative alignment estimates global affine geometry from low-frequency approximations and transfers geometric and contextual cues to high-frequency correlation reasoning for reliability-aware residual refinement. Guided subband fusion jointly aggregates the aligned frequency coefficients under propagated source and alignment guidance, coordinates complementary low- and high-frequency information, and reconstructs the fused image through the inverse wavelet transform. Extensive experiments on RGB-T datasets with real-world and synthetic misalignments demonstrate consistently competitive performance in alignment and fusion, validating the effectiveness of source-anchored guidance for weakly registered RGB-T images.
\end{abstract}

% keywords can be removed

\section{Introduction}
RGB-T image fusion aims to integrate complementary information from visible and thermal sensors into a unified representation \cite{li2026multimodal, li2026hyperbolic}. Visible images provide rich textures and scene details, whereas thermal images emphasize radiation-sensitive targets and remain informative under adverse illumination. Their combination benefits vision tasks, including object detection, semantic segmentation, and scene understanding \cite{zhu2025wavemamba,gao2025visible, wu2025every,zhong2025amdanet}. In practice, however, differences in sensor placement, field of view, and sampling characteristics introduce residual geometric misregistration \cite{li2025bi,hu2025balancing}. Meanwhile, the distinct imaging mechanisms of visible and thermal sensors produce cross-modal appearance discrepancies. Directly combining heterogeneous observations can generate ghosting artifacts, structural blur, and biased information aggregation.

To address these challenges, methods reduce cross-modal discrepancy through a registration-friendly proxy or shared representation before estimating geometric correspondence \cite{wang2022unsupervised,wang2024improving,xu2022rfnet,xu2023murf,jiang2025harmonized,lu2025net,li2026uncertainty}. Approaches couple registration and fusion via feature sharing, fusion feedback, or fusion-derived supervision \cite{wang2024improving,xu2022rfnet,xu2023murf,lu2025net,tang2025c2rf,10856402}. Although these developments improve correspondence learning and task coordination, intermediate representations facilitate registration, whereas fusion information serves as feedback or supervision. Cross-modal information established during discrepancy reduction is rarely retained and jointly exploited for alignment and fusion. Moreover, deformation is estimated from spatial images or mixed latent features through an undifferentiated process, although coarse scene layout and local structural details provide distinct geometric evidence. Consequently, cross-modal representation, alignment, and fusion remain insufficiently coordinated, leaving cross-stage and cross-frequency dependencies underexploited.

These limitations motivate the use of a representation that distinguishes global layout from local structure and supports information transfer across stages. Frequency decomposition is well suited here because it organizes image content according to spatial frequency. Low frequency components retain coarse radiometric variation and global scene layout, making them suitable for global parametric estimation. High frequency components emphasize localized edges and textures, providing complementary evidence for refining local residual displacement after coarse warping. This distinction also applies to fusion because low frequency intensity organization and high frequency structural details require different aggregation behavior. Frequency representations can therefore provide a common basis for cross-modal discrepancy reduction, hierarchical alignment, and adaptive fusion. This shared basis allows information extracted during discrepancy reduction to guide subsequent alignment and fusion stages.

\begin{figure}[t]
\centering
\includegraphics[width=1\columnwidth]{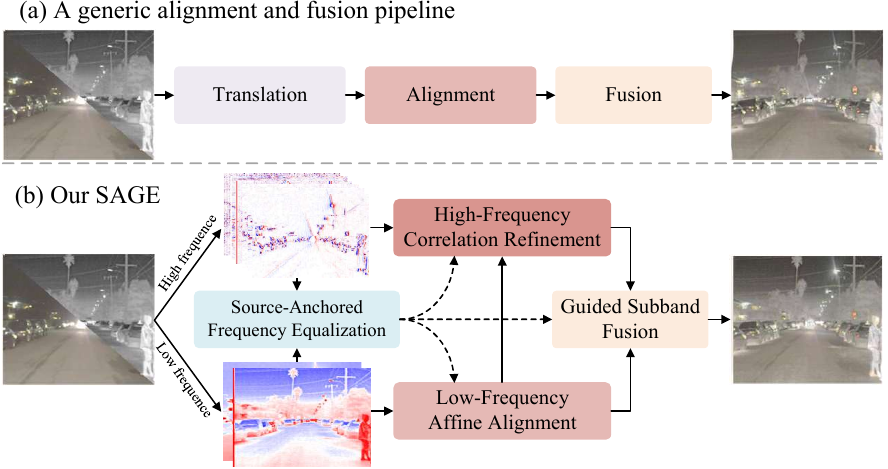}
\caption{Comparison between a conventional pipeline based on image translation and SAGE. (a) Image translation, alignment, and fusion are performed sequentially. (b) SAGE uses source-anchored frequency equalization to guide hierarchical alignment from low to high frequencies and subband fusion. Solid arrows denote feature and dependency flows, while dashed arrows denote guidance propagation.}
\label{fig1}
\end{figure}

Based on this observation, we propose source-anchored guidance via frequency equalization for hierarchical RGB-T alignment and fusion (SAGE), as shown in Figure~\ref{fig1}. Source-anchored frequency equalization (SAFE) jointly models paired wavelet coefficients while retaining source-indexed streams. It applies bounded modulation to low-frequency components, preserves high-frequency structures, and extracts structural and gain cues available to alignment and fusion. The resulting cues propagate information learned during discrepancy reduction to geometric alignment and subband fusion.
Hierarchical frequency collaborative alignment (HFCA) assigns global geometry estimation to low-frequency affine alignment (LFA) and residual local correction to high-frequency correlation refinement (HCR), with LFA transferring affine geometry and low-frequency context to HCR. Guided subband fusion (GSF) then integrates aligned coefficients with propagated source cues and alignment reliability for frequency-specific aggregation and cross-frequency coordination before reconstruction. This organization links source preservation, low-to-high geometric reasoning, and reliability-aware fusion. We examine these design choices through component ablations, alternative frequency-branch assignments, and stage-wise guidance routing.
The main contributions of this work are summarized as follows:
\begin{itemize}
\item We propose a unified source-anchored frequency guidance framework for RGB-T alignment and fusion. It combines invertible joint encoding with source-aware frequency modeling to preserve modality-specific structures and propagate structural and gain cues across stages.

\item We develop a hierarchical frequency collaborative alignment architecture that models the global-to-local dependency between low- and high-frequency information. Low-frequency information establishes global geometric correspondence and provides contextual guidance for high-frequency local refinement.

\item We design a guided subband fusion that combines source-derived cues with correspondence and reliability information for frequency-specific aggregation and cross-frequency interaction. 
% Extensive experiments on multiple RGB-T benchmarks demonstrate superior alignment and fusion performance over state-of-the-art methods.
Extensive experiments demonstrate competitive overall performance across alignment and fusion metrics.
\end{itemize}

\section{Related Work}
\noindent \textbf{Image Alignment.}
Geometric misalignment is a major obstacle to reliable infrared and visible image fusion, since spatial discrepancies can introduce ghosting and structural inconsistency into fused results. Existing methods generally alleviate cross modal appearance discrepancies before estimating geometric correspondence. UMF-CMGR \cite{wang2022unsupervised} and IMF \cite{wang2024improving} progressively model cross modal deformation through transformation and refinement based registration pipelines, while HR4IR \cite{jiang2025harmonized} constructs a harmonized representation domain and alternately searches for cross modal correspondences and geometric transformations. RFNet \cite{xu2022rfnet} and MURF \cite{xu2023murf} further formulate registration and fusion as mutually reinforcing tasks, allowing alignment and fusion to benefit from their reciprocal interaction. Recent studies increasingly couple registration with fusion to strengthen correspondence learning. AU-Net \cite{lu2025net} jointly optimizes the two tasks at the feature level, C2RF \cite{tang2025c2rf} exploits commonality mining and fusion guided contrastive learning, and MulFS-CAP \cite{10856402} introduces fusion supervision to facilitate alignment perception for unregistered inputs. Self supervised correspondence modeling has also received growing attention. B-SR \cite{li2025bi} enforces deformation consistency through bidirectional proxy transformations, whereas Hy-CycleAlign \cite{li2026hyperbolic} combines cyclic registration with hyperbolic correspondence modeling. Despite these advances, existing methods predominantly estimate deformation from spatial images or mixed latent representations, leaving heterogeneous frequency components largely subject to an undifferentiated alignment process. Our method instead performs hierarchical frequency aware alignment, where low frequency affine estimation provides a coarse geometric prior for subsequent high frequency correlation refinement.

\noindent \textbf{Image Fusion.}
Infrared and visible image fusion aims to preserve thermal saliency while retaining visible structural details. Recent studies have increasingly explored adaptive interaction and information selection under diverse conditions \cite{wang2025source, dai2025multi,bai2025task}. DCEvo \cite{liu2025dcevo} improves multimodal integration through discriminative cross dimensional interaction, while EMMA \cite{zhao2024equivariant} introduces equivariant learning for multimodal information integration. TG-ECNet \cite{sun2025task} introduces task aware gating and multi expert collaboration for degraded multimodal fusion, whereas HCLFuse \cite{guo2026revisiting} explores generative fusion through information decomposition and physically guided generation. More recent methods further improve robustness and adaptivity through domain adaptation and region dependent fusion strategies, as demonstrated by DAFusion \cite{guan2026domain} and RegionFuse \cite{xia2026regionfuse}. Despite these advances, fusion decisions are still mainly derived from current source or feature representations. Our method instead uses source anchored structural and gain cues generated before fusion to condition frequency dependent gating and cross frequency interaction, enabling guided information aggregation across different frequency components.

\begin{figure*}[t]
\centering
\includegraphics[width=1.0\textwidth]{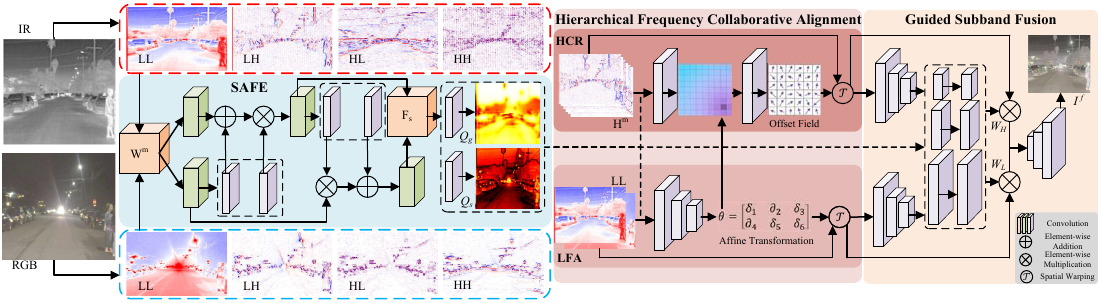}
\caption{Overview of SAGE. The source anchored frequency guidance mechanism consistently propagates structural and gain cues across frequency equalization, hierarchical alignment, and subband fusion. SAFE reduces low frequency discrepancy while preserving source high frequency details. HFCA combines low frequency affine alignment with reliability guided high frequency refinement. GSF adaptively fuses the aligned subbands and reconstructs the final image through inverse Haar wavelet transform.}
\label{fig2}
\end{figure*}

\section{Method}

SAGE is organized around a unified source-anchored frequency guidance mechanism that connects frequency equalization, hierarchical alignment, and subband fusion, as shown in Figure \ref{fig2}. This design preserves the distinct roles of low- and high-frequency information while consistently propagating source-derived cues across processing stages.

\subsection{Overview}
SAGE takes a weakly registered RGB-T pair $(I^{v},I^{t})$ as input and produces a fused image $I^{f}$ through source-anchored frequency equalization, hierarchical alignment, and guided subband fusion. We apply the Haar wavelet transform $\mathcal{D}$ to decompose each input into a low-low (LL) approximation subband and three directional detail subbands: low-high (LH), high-low (HL), and high-high (HH). For $m\in\{v,t\}$,
\begin{equation}
\begin{aligned}
\mathcal{D}\bigl(I^{m}\bigr)
&=\bigl(L_{\mathrm{LL}}^{m},\mathbf{H}^{m}\bigr),\\
\mathbf{H}^{m}
&=\bigl[H_{\mathrm{LH}}^{m},
H_{\mathrm{HL}}^{m},
H_{\mathrm{HH}}^{m}\bigr].
\end{aligned}
\label{eq1}
\end{equation}

We denote the complete wavelet representation as $\mathbf{W}^{m}=[L_{\mathrm{LL}}^{m},\mathbf{H}^{m}]$. An invertible joint encoder transforms $(\mathbf{W}^{v},\mathbf{W}^{t})$ into source-specific joint feature $F_s$. SAFE extracts structural and gain cues while preserving source-specific coefficients. HFCA integrates LFA to estimate affine geometry in the LL band and HCR to correct residual high-frequency misalignment. GSF then performs guidance-conditioned subband aggregation and cross-frequency coordination before inverse wavelet reconstruction.

SAFE performs pair conditioned equalization while preserving the coefficients associated with each source. For $m\in\{v,t\}$, the gain prediction head $\mathcal{P}_{g}^{m}$ processes $\mathbf{W}^{v}$, $\mathbf{W}^{t}$, their elementwise absolute difference, and the joint feature $F_s$. It predicts a bounded residual gain $\Gamma^{m}$ to adaptively modulate the frequency representation while retaining the structural characteristics of the corresponding source.
\begin{equation}
\begin{aligned}
\Gamma^{m}
&=
\gamma\tanh\!\bigl(
\mathcal{P}_{g}^{m}
(\mathbf{W}^{v},\mathbf{W}^{t},
|\mathbf{W}^{v}-\mathbf{W}^{t}|,F_s)
\bigr),\\
\widehat{L}_{\mathrm{LL}}^{m}
&=
L_{\mathrm{LL}}^{m}
\odot
\bigl(1+\Gamma^{m}\bigr),\\
\widehat{\mathbf{H}}^{m}
&=
\mathbf{H}^{m},
\end{aligned}
\label{eq:safe}
\end{equation}
$\odot$ denotes elementwise multiplication, $\gamma$ limits modulation magnitude, and the hat indicates equalized coefficients. Restricting modulation to the LL subband reduces intensity discrepancies across modalities without altering source detail coefficients. SAFE derives a structural cue $Q_s$ and a gain cue $Q_g$ from $F_s$ and propagates both to HFCA and GSF.

\subsection{Hierarchical Frequency Collaborative Alignment (HFCA)}
HFCA models RGB-T alignment through hierarchical frequency collaboration between LFA and HCR. LFA estimates global affine geometry from low-frequency cues, while HCR refines local residual displacement using high-frequency details after compensation. The transferred geometry and low-frequency context enable coordinated use of complementary frequency information across both components.

\noindent \textbf{Low-Frequency Affine Alignment (LFA).}
Conditioned on the structural and gain cues $Q_s$ and $Q_g$, LFA encodes the paired approximation subbands into a contextual representation $F_L$. 
% Global aggregation of $F_L$ predicts an affine transformation $\Theta$, parameterized as a bounded residual around the identity transformation. 
Global aggregation of $F_L$ predicts a $2\times3$ affine transformation $\theta$, parameterized by six bounded residuals $\delta_1,\ldots,\delta_6$ around the identity transformation.
For notational clarity, we present the visible-to-thermal alignment below, taking the thermal modality as the reference. The estimated affine transformation maps the visible coefficients to the thermal coordinate system and provides the initial geometric compensation for the visible high-frequency coefficients. The global alignment is formulated as
\begin{equation}
\begin{aligned}
L_{a}^{v}
&=
\mathcal{T}_{\theta}
\bigl(\widehat{L}_{\mathrm{LL}}^{v}\bigr),
&
L_{a}^{t}
&=
\widehat{L}_{\mathrm{LL}}^{t},
\\
\overline{\mathbf{H}}^{v}
&=
\mathcal{T}_{\theta}
\bigl(\widehat{\mathbf{H}}^{v}\bigr),
&
\overline{\mathbf{H}}^{t}
&=
\widehat{\mathbf{H}}^{t},
\end{aligned}
\label{eq:coarse_alignment}
\end{equation}
$\mathcal{T}_{\theta}$ denotes affine warping, and the overline indicates high-frequency coefficients after global compensation.

\noindent \textbf{High-Frequency Correlation Refinement (HCR).}
HCR constructs a normalized local correlation volume from the globally compensated high-frequency features over a restricted search neighborhood. Combined with transferred low-frequency context and source-derived cues, the correlation evidence estimates a residual offset field $D$. HCR further derives a structural reliability map $R\in[0,1]$ from coherent high-frequency responses and low-frequency gradients. The reliability-gated residual deformation and aligned high-frequency coefficients are computed as
\begin{equation}
\begin{aligned}
\widetilde{D}
&=
R\odot D,\\
\mathbf{H}_{a}^{v}
&=
\mathcal{T}_{\widetilde{D}}
\bigl(\overline{\mathbf{H}}^{v}\bigr),
&
\mathbf{H}_{a}^{t}
&=
\overline{\mathbf{H}}^{t}.
\end{aligned}
\label{eq:residual_alignment}
\end{equation}

Through LFA and HCR, HFCA establishes a frequency-directed geometric dependency in which source-derived cues and structural reliability jointly govern a robust, adaptive residual displacement recovery process.

\subsection{Guided Subband Fusion (GSF)}
GSF incorporates guidance from SAFE and HFCA into frequency-specific fusion. The low-frequency gate is estimated from aligned approximation coefficients, $Q_s$, and $Q_g$, while the high-frequency gate also incorporates local correlation responses and the reliability map. Gates $W_L$ and $W_H$ are applied to aligned low- and high-frequency coefficients, respectively. The initial fused coefficients are computed as
\begin{equation}
\begin{aligned}
\widetilde{L}^{f}
&=
W_{L}\odot L_{a}^{v}
+
\bigl(1-W_{L}\bigr)\odot L_{a}^{t},
\\
\widetilde{\mathbf{H}}^{f}
&=
W_{H}\odot\mathbf{H}_{a}^{v}
+
\bigl(1-W_{H}\bigr)\odot\mathbf{H}_{a}^{t}.
\end{aligned}
\label{eq:subband_fusion}
\end{equation}

The frequency-specific gates integrate global intensity organization and local structural evidence into the initial subband fusion. The fused coefficients are subsequently refined through cross-frequency interaction, with the reliability map regulating the high-frequency update according to local correspondence confidence. The final coefficients $L^{f}$ and $\mathbf{H}^{f}$ are reconstructed as $I^{f}=\mathcal{D}^{-1}(L^{f},\mathbf{H}^{f})$.

\subsection{Loss Function}
SAGE is trained on weakly registered RGB-T image pairs. Its loss comprises equalization, alignment, and fusion terms. The equalization term $\mathcal{L}_{\mathrm{eq}}$ applies Charbonnier loss \cite{9577298} to invertible reconstruction and source-consistency errors, with predicted gain-field regularization. The alignment term $\mathcal{L}_{\mathrm{align}}$ enforces correspondence between aligned low-frequency approximations and high-frequency energy representations via normalized cross-correlation \cite{xu2022rfnet}, while regularizing affine transformation and residual offsets. The fusion term $\mathcal{L}_{\mathrm{fus}}$ comprises maximum-intensity, maximum-gradient, and edge-preservation constraints derived from aligned source images \cite{zhao2023cddfuse}. The overall training loss is defined as
\begin{equation}
\mathcal{L}
=
\lambda_{\mathrm{eq}}\mathcal{L}_{\mathrm{eq}}
+
\lambda_{\mathrm{align}}\mathcal{L}_{\mathrm{align}}
+
\lambda_{\mathrm{fus}}\mathcal{L}_{\mathrm{fus}},
\label{eq:total_loss}
\end{equation}
where $\lambda_{\mathrm{eq}}$, $\lambda_{\mathrm{align}}$, and $\lambda_{\mathrm{fus}}$ denote the weighting coefficients of the corresponding loss terms.

\section{Experiments}
\subsection{Setup}
\noindent \textbf{Datasets.}
We evaluate SAGE on three public RGB-T datasets, including DroneVehicle \cite{sun2022drone}, MFNet \cite{ha2017mfnet}, and RoadScene \cite{xu2020aaai}. DroneVehicle contains aerial RGB-T pairs with inherent cross-modal misalignment arising from platform motion, sensor parallax, and viewpoint variation. Its original image pairs are used without additional geometric perturbation. MFNet and RoadScene comprise ground-level traffic scenes captured under diverse illumination conditions.

\noindent \textbf{Misalignment Settings.}
We evaluate DroneVehicle, MFNet, and RoadScene under complementary real-world, non-rigid, and rigid RGB-T misalignment scenarios. DroneVehicle retains its inherent aerial-view misalignment without additional perturbation. 
On MFNet, thermal images serve as references, while visible images undergo translations and elastic deformations using Gaussian kernels of sizes $85$--$101$ with standard deviations of $24$--$32$ pixels.
On RoadScene, visible images serve as references, while thermal images are shifted leftward by $0.5\%$ to $1.5\%$ of image width. These settings cover both modality directions under real, synthetic non-rigid, and rigid conditions.

\noindent \textbf{Metrics.}
We report five metrics to evaluate both fusion quality and cross-modal alignment accuracy. Average gradient (AG) and spatial frequency (SF) characterize the preservation of local details and spatial activity in the fused image. The sum of correlations of differences (SCD) assesses complementary information transfer and overall consistency with the source observations. To specifically evaluate alignment performance, we further report the 95th-percentile Hausdorff distance (HD95) and average symmetric surface distance (ASSD), which measure the geometric discrepancy between corresponding structural boundaries after alignment. 
Higher AG, SF, and SCD values indicate better fusion quality, while lower HD95 and ASSD values indicate more accurate structural alignment.

\noindent \textbf{Implementation details.}
SAGE is implemented and trained on an NVIDIA GeForce RTX 3090. The network is optimized for 3000 epochs using AdamW with an initial learning rate of $5\times10^{-5}$. The frequency equalization parameter is set to $\gamma=0.2$, while the loss weights are set to $\lambda_{\mathrm{eq}}=0.6$, $\lambda_{\mathrm{align}}=1$, and $\lambda_{\mathrm{fus}}=0.8$.

\noindent\textbf{Quantitative evaluation.}
\noindent\textit{Alignment quality.}
As shown in Table~\ref{tab:quantitative_comparison}, SAGE demonstrates competitive alignment performance across three datasets, which cover aerial observations and spatially varying and rigid deformations. Its advantage is most evident on MFNet, where local displacement varies across the image. The results indicate that SAGE accommodates deformation patterns while maintaining structural correspondence. This consistency is in line with the use of source-anchored frequency cues and frequency-aware global and local correspondence estimation.

\noindent\textit{Fusion quality.}
SAGE demonstrates competitive overall fusion performance across three benchmarks while preserving fine structures and complementary visible and thermal information. Its strong RoadScene results and generally stable performance elsewhere suggest a favorable balance among detail preservation structural fidelity and cross modal information retention under diverse imaging conditions. This balance arises from the complete SAGE pipeline where source frequency information is retained during equalization geometric correspondence is refined hierarchically and aligned content is aggregated through guided subband fusion.

\subsection{Comparing with SOTA Methods}
We compare SAGE against eight representative state-of-the-art methods developed for misaligned RGB-T image fusion. The selected approaches account for geometric inconsistency through correction, explicit registration, or joint registration and fusion. They include SuperFusion \cite{TANG2022SuperFusion}, ReCoNet \cite{huang2022reconet}, MURF \cite{xu2023murf}, UMF-CMGR \cite{wang2022unsupervised}, IMF \cite{wang2024improving}, AU-Net \cite{lu2025net}, C2RF \cite{tang2025c2rf}, and FusionRegister \cite{bian2026fusionregister}. 

\begin{figure*}[t]
    \centering
    \includegraphics[width=1.0\textwidth]{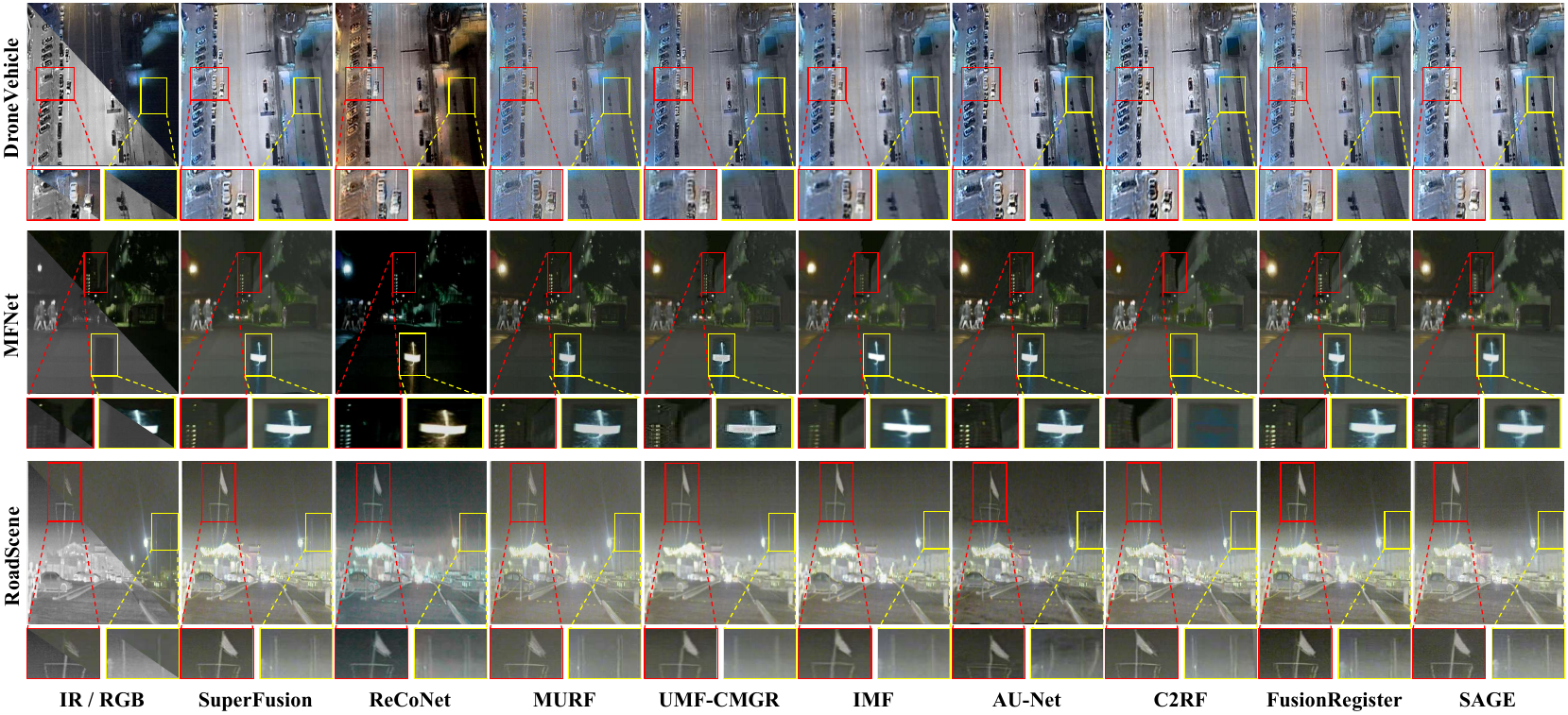}
    \caption{Qualitative comparisons on DroneVehicle, MFNet, and RoadScene.}
    \label{result_all}
\end{figure*}

\noindent \textbf{Qualitative evaluation.}
\noindent\textit{Alignment quality.}
As shown in Figure~\ref{result_all}, SAGE achieves sharper, spatially coherent alignment across the three datasets. On DroneVehicle, it preserves separable contours among densely parked vehicles and accurately localizes the isolated road target, while competing methods produce blurred boundaries or duplicated structures. On MFNet, SAGE maintains compact contours around the bright target and consistent facade patterns despite spatially varying displacement. On RoadScene, it reconstructs the flag and pole structures with clear single edges and minimal ghosting. These results indicate that SAGE handles both local and spatially varying misalignment.

\noindent\textit{Fusion quality.}
SAGE also provides balanced integration of infrared saliency and visible details. On DroneVehicle, it preserves vehicle structures, the isolated target, and road context without noticeable distortion. On MFNet, salient thermal targets are retained with tree and facade textures. On RoadScene, the flag boundary, roadside structures, and lane markings remain clear without excessive brightness spreading. Overall, SAGE produces fewer ghosting artifacts and less source bias by adaptively regulating frequency-specific aggregation according to alignment reliability.

\begin{table*}[t]
    \centering
    \LARGE
    \resizebox{\linewidth}{!}{
        % \begin{tabular}{l|ccccc|ccccc|ccccc|ccccc}
        \begin{tabular}{l|ccccc|ccccc|ccccc}
            \toprule
            \multirow{2}{*}{Method}
            & \multicolumn{5}{c|}{\textbf{DroneVehicle}}
            & \multicolumn{5}{c|}{\textbf{MFNet}}
            % & \multicolumn{5}{c}{\textbf{LLVIP}}
            & \multicolumn{5}{c}{\textbf{RoadScene}}\\
            \cmidrule(lr){2-6}
            \cmidrule(lr){7-11}
            \cmidrule(lr){12-16}
            % \cmidrule(lr){17-21}
            & AG $\uparrow$ & SCD $\uparrow$ & SF $\uparrow$ & HD95 $\downarrow$ & ASSD $\downarrow$
            & AG $\uparrow$ & SCD $\uparrow$ & SF $\uparrow$ & HD95 $\downarrow$ & ASSD $\downarrow$
            % & AG $\uparrow$ & SCD $\uparrow$ & SF $\uparrow$ & HD95 $\downarrow$ & ASSD $\downarrow$
            & AG $\uparrow$ & SCD $\uparrow$ & SF $\uparrow$ & HD95 $\downarrow$ & ASSD $\downarrow$\\
            \midrule
            SuperFusion
            & $6.11$ & $0.91$ & $17.50$ & $\underline{30.12}$ & $\underline{7.55}$
            & $2.96$ & $1.06$ & $7.90$ & $63.83$ & $14.49$
            % & $4.18$ & $1.25$ & $14.10$ & $120.91$ & $29.97$
            & $4.33$ & $1.21$ & $11.67$ & $44.43$ & $\underline{10.38}$ \\
            ReCoNet
            & $5.20$ & $\mathbf{1.48}$ & $13.60$ & $30.53$ & $7.71$
            & $\underline{3.15}$ & $\mathbf{1.53}$ & $9.51$ & $69.72$ & $17.71$
            % & $3.64$ & $1.37$ & $11.43$ & $117.57$ & $29.30$
            & $3.65$ & $\underline{1.27}$ & $9.02$ & $45.72$ & $11.57$ \\
            MURF
            & $2.34$ & $0.89$ & $5.72$ & $44.95$ & $12.06$
            & $3.02$ & $\underline{1.36}$ & $\underline{10.47}$ & $\underline{30.47}$ & $\underline{6.77}$
            % & $6.65$ & $1.24$ & $19.69$ & $88.25$ & $20.64$
            & $3.49$ & $\underline{1.27}$ & $9.79$ & $\underline{39.36}$ & $\mathbf{8.43}$ \\
            UMF-CMGR
            & $4.28$ & $1.02$ & $10.96$ & $31.73$ & $8.93$
            & $1.81$ & $1.07$ & $5.25$ & $100.17$ & $39.25$
            % & $1.76$ & $1.09$ & $4.64$ & $140.75$ & $39.17$
            & $2.90$ & $0.61$ & $3.86$ & $65.27$ & $22.19$ \\
            IMF
            & $4.11$ & $1.01$ & $8.44$ & $30.97$ & $8.62$
            & $1.23$ & $0.77$ & $3.77$ & $68.03$ & $25.70$
            % & $1.82$ & $1.07$ & $4.82$ & $119.41$ & $36.29$
            & $3.50$ & $1.25$ & $7.43$ & $56.76$ & $18.32$ \\
            AU-Net
            & $\underline{6.15}$ & $1.01$ & $18.16$ & $31.34$ & $7.76$
            & $2.96$ & $1.22$ & $9.05$ & $77.34$ & $20.34$
            % & $4.31$ & $1.05$ & $15.94$ & $82.34$ & $21.51$
            & $3.77$ & $1.26$ & $9.68$ & $43.34$ & $10.47$ \\
            C2RF
            & $5.66$ & $0.82$ & $\underline{18.42}$ & $35.24$ & $9.63$
            & $1.63$ & $0.41$ & $4.48$ & $92.36$ & $24.41$
            % & $4.65$ & $0.80$ & $15.09$ & $49.60$ & $10.64$
            & $\underline{4.56}$ & $1.19$ & $\underline{13.42}$ & $44.06$ & $11.10$ \\
            FusionRegister
            & $5.54$ & $1.08$ & $14.95$ & $31.10$ & $7.72$
            & $3.10$ & $1.04$ & $8.87$ & $30.76$ & $12.74$
            % & $4.65$ & $0.80$ & $15.09$ & $49.60$ & $10.64$
            & $3.88$ & $1.22$ & $10.70$ & $51.13$ & $20.16$\\
            SAGE
            & $\mathbf{6.18}$ & $\underline{1.13}$ & $\mathbf{18.73}$ & $\mathbf{29.19}$ & $\mathbf{7.42}$
            & $\mathbf{3.17}$ & $1.28$ & $\mathbf{11.49}$ & $\mathbf{21.65}$ & $\mathbf{4.24}$
            % & $-$ & $-$ & $-$ & $-$ & $-$
            & $\mathbf{5.18}$ & $\mathbf{1.28}$ & $\mathbf{15.05}$ & $\mathbf{38.52}$ & $14.28$ \\
            \bottomrule
        \end{tabular}%
    }
    \caption{Quantitative comparison with SOTA methods. The \textbf{bold}/\underline{underline} indicates the best and runner-up.}
    \label{tab:quantitative_comparison}
\end{table*}

\begin{figure}[t]
\centering
\includegraphics[width=.6\columnwidth]{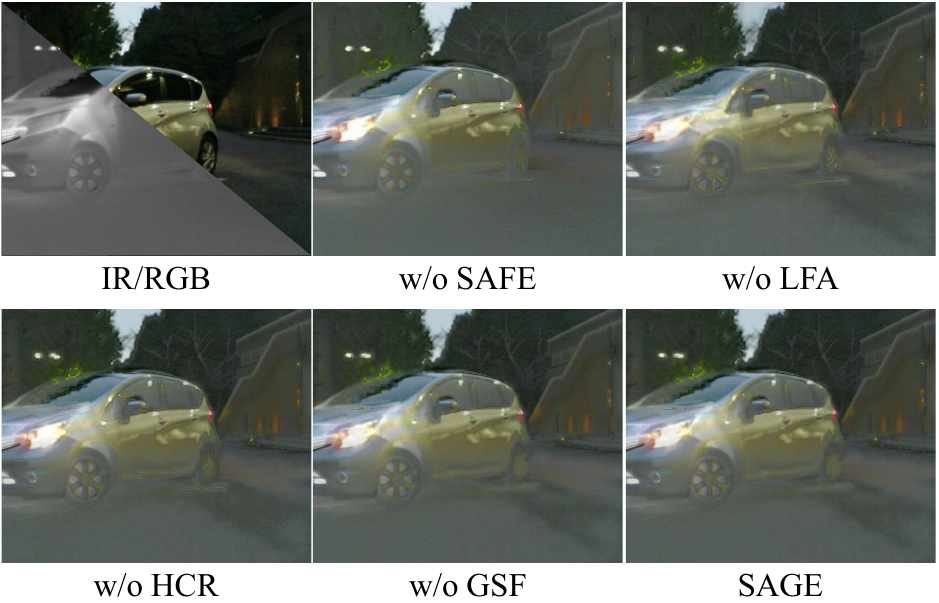}
\caption{Qualitative ablation results on MFNet.}
% \caption{The analysis of the ablation experiment was conducted using the MFNet dataset.}
\label{result_ab}
\end{figure}

\begin{table}[t]
\centering
\resizebox{\linewidth}{!}{
% \tiny
% \begin{tabular}{c|c c c c c|c c c c c}
\begin{tabular}{c|c c c c c}
\toprule
Methods &AG $\uparrow$ & SCD $\uparrow$ & SF $\uparrow$ & HD95 $\downarrow$ & ASSD $\downarrow$ \\
% Methods &$SAFE$ &$LFA$ &$HCR$ &$HFCA$ &$GSF$ &AG $\uparrow$ & SCD $\uparrow$ & SF $\uparrow$ & HD95 $\downarrow$ & ASSD $\downarrow$ \\
\midrule
w/o SAFE &$2.72$ &$1.19$  &$9.86$ &$29.46$ &$5.82$ \\
% w/o LFA &$-$&$-$ &$-$  &$-$ &$-$ \\
w/o LFA &$3.09$&$\mathbf{1.36}$ &$7.97$  &$35.40$ &$7.09$ \\
w/o HCR &$2.68$&$1.14$ &$8.58$  &$36.20$ &$7.36$ \\
w/o GSF &$2.76$&$1.35$ &$8.11$  &$35.08$ &$6.85$ \\
% w/o CFI &$-$&$-$ &$-$  &$-$ &$-$ \\
\midrule
SAGE &$\mathbf{3.17}$ &$1.28$ &$\mathbf{11.49}$ &$\mathbf{21.65}$ &$\mathbf{4.24}$ \\
\bottomrule
\end{tabular}
}
% \caption{Ablation experiment results in MFNet dataset. LFA denotes low frequency affine alignment, HCR denotes high frequency correlation refinement, and GSF denotes guided subband fusion.}
\caption{Ablation results on MFNet. LFA, HCR, and GSF denote low-frequency affine alignment, high-frequency correlation refinement, and guided subband fusion, respectively.}
\label{table_abe}
\end{table}

\subsection{Ablation Studies}

We conduct ablation studies on MFNet to examine source-aware frequency equalization, hierarchical frequency collaborative alignment, and guided subband fusion, with Table~\ref{table_abe} demonstrating the competitive overall performance of the complete SAGE model. The two alignment stages are removed separately to distinguish global affine correction from high-frequency residual refinement. Removing source-aware guidance weakens structural preservation and source consistency. Disabling either alignment stage introduces greater misalignment, confirming the complementary roles of global geometry estimation and local residual correction. Removing guided subband fusion reduces detail preservation and disrupts the coordinated aggregation of aligned frequency information.
The visual comparisons in Figure~\ref{result_ab} support these findings. Variants without either alignment stage exhibit more visible ghosting and contour displacement around the vehicle. Removing source-aware guidance or guided fusion produces blurrier details and less balanced source integration. In contrast, the complete model preserves sharper boundaries, more coherent structures, and a better balance between thermal saliency and visible appearance. These results validate the coordinated contributions of frequency guidance, hierarchical alignment, and subband fusion.

\begin{figure}[t]
\centering
\includegraphics[width=1.\columnwidth]{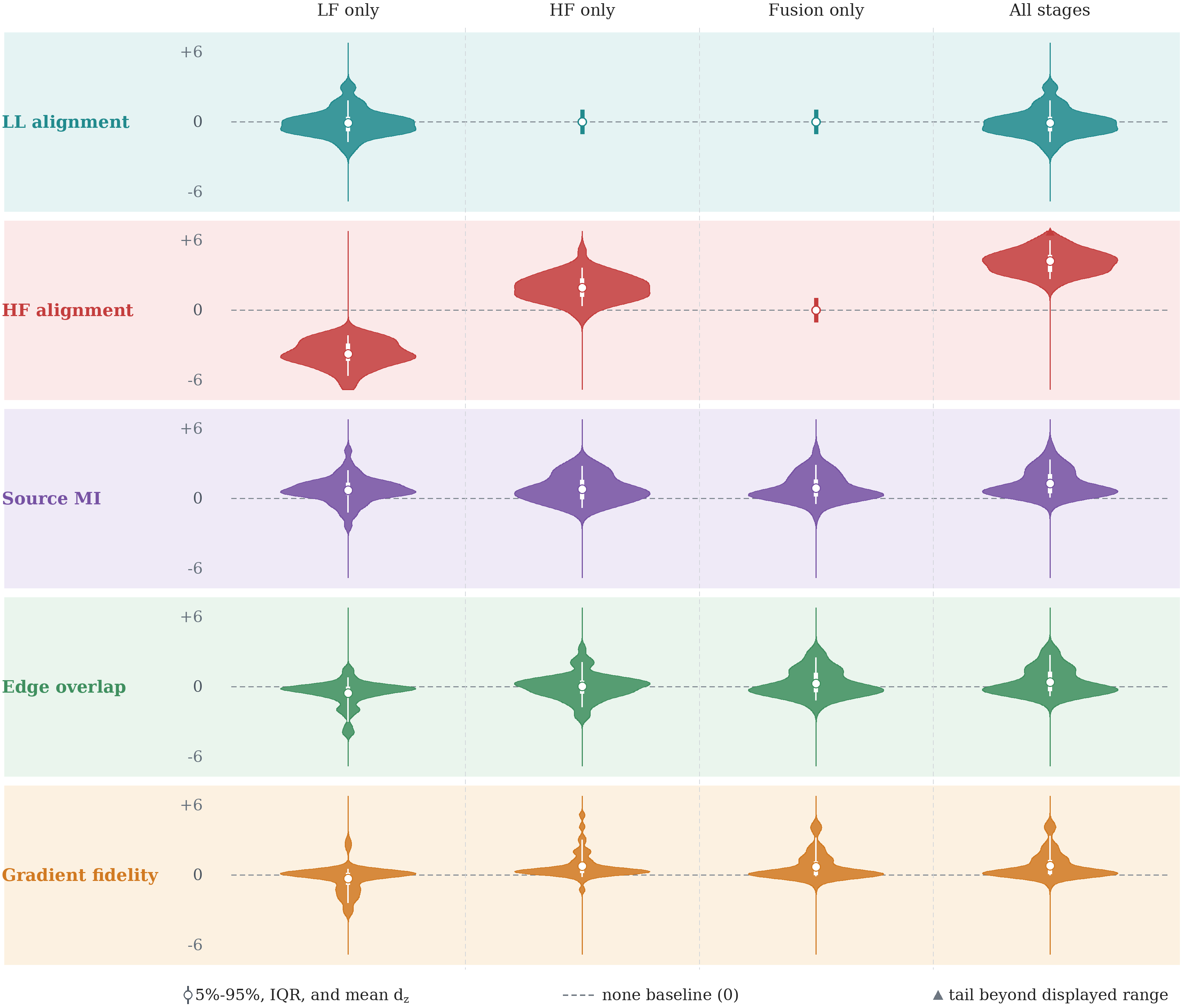}
\caption{Effects of guidance propagation to low-frequency alignment, high-frequency alignment, fusion, or all stages.}
% Violins show paired standardized effect sizes relative to the no-guidance baseline, with positive values indicating improvement.}
\label{fig:guidance_routing}
\end{figure}
\subsection{Effect of Guidance Propagation}
Figure~\ref{fig:guidance_routing} compares guidance delivered to low frequency alignment, high frequency alignment, fusion, and all stages with the no guidance reference. Using identical test pairs and misalignment conditions, we vary only the recipient stages. For each metric, the paired difference $\delta_i$ is oriented so that positive values indicate improvement and standardized by its cross-sample deviation as $d_z=\overline{\delta}/s_{\delta}$. 

\noindent\textit{Alignment.}
Low frequency guidance produces a broad LL response but degrades HF alignment, while high frequency guidance improves the HF response and fusion guidance leaves both measures near the baseline. Guidance across all stages yields the strongest positive HF response while maintaining the LL response near the reference, supporting the coarse to fine dependency in HFCA.

\noindent\textit{Fusion.}
Fusion guidance improves source mutual information, edge overlap, and gradient fidelity without materially changing alignment, while high frequency guidance also benefits these measures and low frequency guidance degrades the edge and gradient responses. Guidance across all stages maintains positive mean effects for all three indicators together with the strongest HF response, supporting coordinated propagation throughout SAGE.

\begin{figure}[t]
    \centering
    \begin{subfigure}[t]{0.49\textwidth}
        \centering
        \includegraphics[width=\linewidth]
        {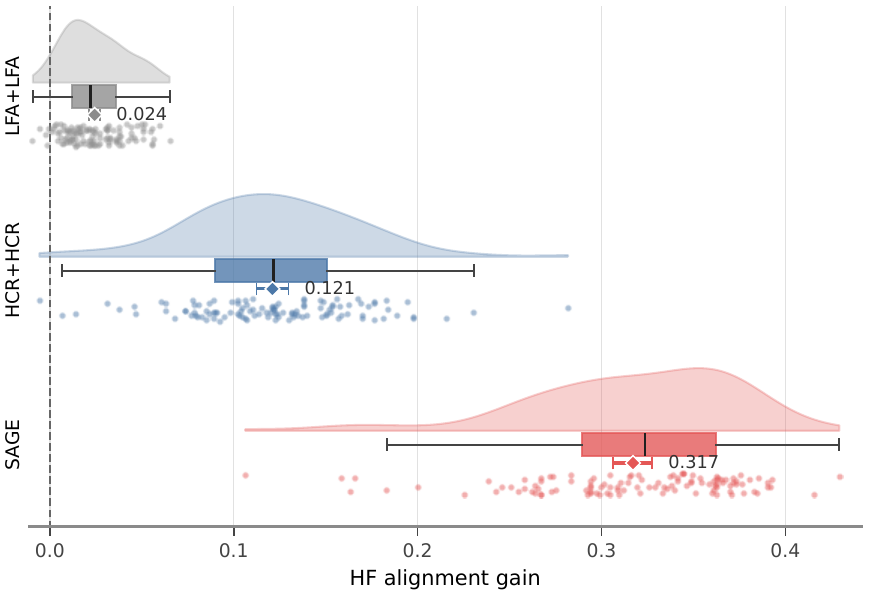}
        \caption{High-frequency alignment improvement.}
        \label{fig:guidance_routing_hf_gain}
    \end{subfigure}
    \hfill
    \begin{subfigure}[t]{0.485\textwidth}
        \centering
        \includegraphics[width=\linewidth]
        {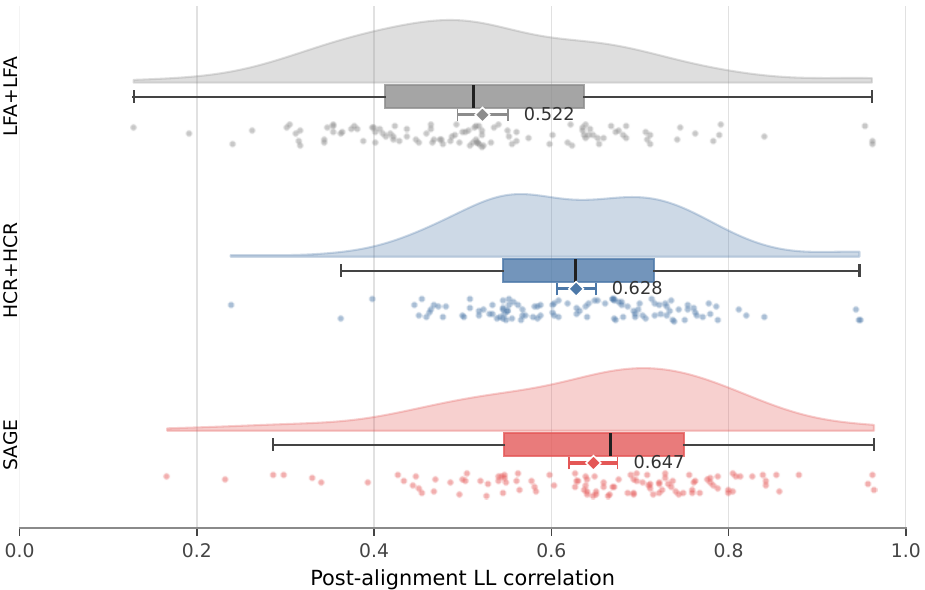}
        \caption{Low-frequency correspondence preservation.}
        \label{fig:guidance_routing_ll_corr}
    \end{subfigure}
\caption{
Comparison of collaboration strategies in (a) HF alignment gain and (b) post-alignment LL correlation.
}
    \label{fig:hierarchical_alignment}
\end{figure}

\subsection{Frequency Specialized Hierarchical Collaboration}
Low frequency components mainly encode global scene structure, whereas high frequency components preserve edges and local details that are more sensitive to spatial misalignment. We therefore compare three alignment configurations under the same evaluation setting. LFA+LFA applies global affine alignment to both frequency branches, HCR+HCR applies local correlation refinement to both branches, and SAGE assigns LFA to the low frequency branch and HCR to the high frequency branch. As shown in Figure~\ref{fig:hierarchical_alignment}, SAGE achieves the strongest high frequency alignment improvement while preserving the highest low frequency correspondence. LFA+LFA provides limited improvement because a global transformation cannot adequately correct spatially varying high frequency displacement. HCR+HCR performs better but requires local refinement to compensate for both global and local misalignment. In contrast, SAGE first establishes reliable global correspondence in the low frequency branch and then concentrates high frequency refinement on residual local errors. These results demonstrate that frequency specific branch assignment provides more effective and coherent alignment than applying a single mechanism to both frequency components.

\begin{table}[t]
\centering
\resizebox{\linewidth}{!}{
\tiny
\begin{tabular}{c|c c c}
\toprule
Methods & Recall $\uparrow$ & Precision $\uparrow$ & mAP@$0.5$ $\uparrow$ \\
\midrule
SuperFusion     & $23.40$ & $14.11$ & $11.43$ \\
ReCoNet         & $24.59$ & $13.28$ & $10.31$ \\
MURF            & $\underline{27.51}$ & $11.17$ & $10.69$ \\
UMF-CMGR        & $26.94$ & $11.96$ & $10.80$ \\
IMF             & $21.85$ & $11.67$ & $8.83$ \\
% MuIFS           & $-$    & $-$    & $-$ \\
AU-Net          & $24.87$ & $12.54$ & $10.56$ \\
C2RF            & $18.62$ & $10.06$ & $9.38$ \\
FusionRegister  & $23.51$ & $\underline{16.23}$ & $\underline{12.00}$ \\
SAGE   & $\mathbf{33.63}$ & $\mathbf{19.31}$ & $\mathbf{16.04}$ \\
\bottomrule
\end{tabular}
}
\caption{Quantitative object detection results on DroneVehicle. The \textbf{bold}/\underline{underline} indicates the best and runner-up.}
\label{table_det}
\end{table}

\subsection{Downstream Task Evaluation}

To evaluate the downstream benefits of alignment and fusion, we train and test YOLO11m \cite{yolo11_ultralytics} on the unaligned DroneVehicle dataset using fusion results from different methods. Recall, precision, and mAP@$0.5$ are used to assess target coverage, prediction reliability, and detection accuracy, respectively. As reported in Table~\ref{table_det}, SAGE achieves the best performance across all metrics. The simultaneous improvements in recall and precision indicate that SAGE recovers more targets without sacrificing prediction reliability. Qualitative results in the supplementary material further show that SAGE improves the detection of small and densely distributed vehicles and reduces missed and inaccurate detections, demonstrating its effectiveness in preserving structural correspondence and complementary multimodal information for downstream detection.

\subsection{Complexity Comparison}

Table \ref{table_complexity} compares the computational complexity and inference efficiency of SAGE with representative methods. SAGE requires $91.45$G FLOPs and $3.53$M parameters, maintaining moderate computational cost among the compared approaches. Despite jointly performing frequency equalization, hierarchical alignment, and subband fusion, SAGE achieves a runtime of $0.02$s and $56.50$ FPS, demonstrating competitive efficiency. These results indicate a favorable balance between model complexity and inference performance.

\begin{table}[t]
\centering
\resizebox{\linewidth}{!}{
\Large
\begin{tabular}{c|cccc}
\toprule
Methods & FLOPs (G) & Parameters (M) & Time (s) & FPS \\
\midrule
SuperFusion & $16.36$ & $1.96$ & $0.03$ & $38.69$ \\
ReCoNet & $15.33$ & $0.21$ & $0.33$ & $3.05$ \\
MURF & $100.10$ & $4.08$ & $0.12$ & $8.13$ \\
UMF-CMGR & $131.38$ & $14.44$ & $0.09$ & $10.69$ \\
IMF & $123.30$ & $15.70$ & $0.06$ & $17.07$ \\
AU-Net & $14.04$ & $2.27$ & $0.02$ & $60.93$ \\
C2RF & $165.12$ & $2.70$ & $0.06$ & $16.69$ \\
FusionRegister & $49.79$ & $0.77$ & $0.08$ & $12.93$ \\
SAGE & $91.45$ & $3.53$ & $0.02$ & $56.50$ \\
\bottomrule
\end{tabular}
}
% \caption{Comparison of model complexity and inference efficiency across different methods.}
\caption{Model complexity and inference efficiency.}
\label{table_complexity}
\end{table}

\section{Conclusion}
We presented SAGE, a unified frequency-domain framework for weakly registered RGB-T alignment and fusion. SAFE preserves source content and derives structural and gain guidance, HFCA couples global geometry with local residual refinement, and GSF integrates propagated cues and alignment reliability for subband fusion. Extensive experiments demonstrate robust performance across diverse scenes and deformation conditions. Guidance propagation and downstream detection results further confirm that SAGE maintains structural correspondence and task-relevant cross-modal information. Overall, SAGE effectively coordinates source preservation, geometric alignment, and frequency-specific fusion within a unified framework.

% \section*{Acknowledgments}
% This was was supported in part by......

%Bibliography
\bibliographystyle{unsrt}  
\bibliography{references}

@article{li2026multimodal,
  title={Multimodal Image Registration for Low Altitude Platforms: Methods, Challenges, and Future Trends},
  author={Li, Timing and Cao, Bing and Zhu, Pengfei and Li, Kewen},
  journal={Tsinghua Science and Technology},
  year={2026},
  publisher={清华大学出版社}
}

@article{li2026hyperbolic,
  title={Hyperbolic Cycle Alignment for Infrared-Visible Image Fusion},
  author={Li, Timing and Cao, Bing and Feng, Jiahe and Cao, Haifang and Hu, Qinghua and Zhu, Pengfei},
  journal={IEEE Transactions on Image Processing},
  year={2026},
  publisher={IEEE}
}

@article{li2025bi,
  title={Bi-directional Self-Registration for Misaligned Infrared-Visible Image Fusion},
  author={Li, Timing and Cao, Bing and Zhu, Pengfei and Xiao, Bin and Hu, Qinghua},
  journal={arXiv preprint arXiv:2505.06920},
  year={2025}
}

@article{gao2025visible,
  title={Visible-infrared image alignment for UAVs: Benchmark and new baseline},
  author={Gao, Zhinan and Li, Dongdong and Kuai, Yangliu and Chen, Rui and Wen, Gongjian},
  journal={IEEE Transactions on Geoscience and Remote Sensing},
  volume={63},
  pages={1--14},
  year={2025},
  publisher={IEEE}
}

@article{tang2025c2rf,
  title={C2RF: Bridging multi-modal image registration and fusion via commonality mining and contrastive learning},
  author={Tang, Linfeng and Yan, Qinglong and Xiang, Xinyu and Fang, Leyuan and Ma, Jiayi},
  journal={International journal of computer vision},
  volume={133},
  number={8},
  pages={5262--5280},
  year={2025},
  publisher={Springer}
}

@article{lu2025net,
  title={AU-Net: Adaptive unified network for joint multi-modal image registration and fusion},
  author={Lu, Ming and Jiang, Min and Tao, Xuefeng and Kong, Jun},
  journal={IEEE Transactions on Image Processing},
  year={2025},
  publisher={IEEE}
}

@article{wang2022unsupervised,
  title={Unsupervised misaligned infrared and visible image fusion via cross-modality image generation and registration},
  author={Wang, Di and Liu, Jinyuan and Fan, Xin and Liu, Risheng},
  journal={arXiv preprint arXiv:2205.11876},
  year={2022}
}

@article{wang2024improving,
  title={Improving misaligned multi-modality image fusion with one-stage progressive dense registration},
  author={Wang, Di and Liu, Jinyuan and Ma, Long and Liu, Risheng and Fan, Xin},
  journal={IEEE Transactions on Circuits and Systems for Video Technology},
  volume={34},
  number={11},
  pages={10944--10958},
  year={2024},
  publisher={IEEE}
}

@article{jiang2025harmonized,
  title={Harmonized domain enabled alternate search for infrared and visible image alignment},
  author={Jiang, Zhiying and Zhang, Zengxi and Liu, Jinyuan},
  journal={IEEE Transactions on Image Processing},
  year={2025},
  publisher={IEEE}
}

@ARTICLE{10856402,
  author={Li, Huafeng and Yang, Zengyi and Zhang, Yafei and Jia, Wei and Yu, Zhengtao and Liu, Yu},
  journal={IEEE Transactions on Pattern Analysis and Machine Intelligence}, 
  title={MulFS-CAP: Multimodal Fusion-Supervised Cross-Modality Alignment Perception for Unregistered Infrared-Visible Image Fusion}, 
  year={2025},
  volume={47},
  number={5},
  pages={3673-3690},
  doi={10.1109/TPAMI.2025.3535617}}

@inproceedings{liu2025dcevo,
  title={Dcevo: Discriminative cross-dimensional evolutionary learning for infrared and visible image fusion},
  author={Liu, Jinyuan and Zhang, Bowei and Mei, Qingyun and Li, Xingyuan and Zou, Yang and Jiang, Zhiying and Ma, Long and Liu, Risheng and Fan, Xin},
  booktitle={Proceedings of the IEEE/CVF Conference on Computer Vision and Pattern Recognition},
  pages={2226--2235},
  year={2025}
}

@inproceedings{xia2026regionfuse,
  title={RegionFuse: Region-Adaptive Pixel Distribution Learning for Infrared and Visible Image Fusion},
  author={Xia, Jianghan and Song, Hong and Li, Jinfu and Lin, Yucong and Ma, Shihan and Fan, Jingfan and Ai, Danni and Fu, Tianyu and Xiao, Deqiang and Yang, Jian},
  booktitle={Proceedings of the IEEE/CVF Conference on Computer Vision and Pattern Recognition},
  pages={19539--19548},
  year={2026}
}

@inproceedings{guan2026domain,
  title={Domain adaptation guided infrared and visible image fusion},
  author={Guan, Tianwei and Wei, Haozhen and Zhou, Yuhan and Ma, Jun and Xu, Zecheng and Jiang, Zhiying and Liu, Jinyuan and Li, Xingyuan},
  booktitle={Proceedings of the AAAI Conference on Artificial Intelligence},
  pages={4376--4384},
  year={2026}
}

@inproceedings{zhao2024equivariant,
  title={Equivariant multi-modality image fusion},
  author={Zhao, Zixiang and Bai, Haowen and Zhang, Jiangshe and Zhang, Yulun and Zhang, Kai and Xu, Shuang and Chen, Dongdong and Timofte, Radu and Van Gool, Luc},
  booktitle={Proceedings of the IEEE/CVF conference on computer vision and pattern recognition},
  pages={25912--25921},
  year={2024}
}

@inproceedings{sun2025task,
  title={Task-gated multi-expert collaboration network for degraded multi-modal image fusion},
  author={Sun, Yiming and Li, Xin and Zhu, Pengfei and Hu, Qinghua and Ren, Dongwei and Xu, Huiying and Zhu, Xinzhong},
  booktitle={International Conference on Machine Learning},
  pages={57571--57586},
  year={2025},
  organization={PMLR}
}

@article{guo2026revisiting,
  title={Revisiting generative infrared and visible image fusion based on human cognitive laws},
  author={Guo, Lin and Luo, Xiaoqing and Xie, Wei and Zhang, Zhancheng and Li, Hui and Wang, Rui and Feng, Zhenhua and Song, Xiaoning},
  journal={Advances in Neural Information Processing Systems},
  volume={38},
  pages={96322--96352},
  year={2026}
}

@inproceedings{xu2022rfnet,
  title={Rfnet: Unsupervised network for mutually reinforcing multi-modal image registration and fusion},
  author={Xu, Han and Ma, Jiayi and Yuan, Jiteng and Le, Zhuliang and Liu, Wei},
  booktitle={Proceedings of the IEEE/CVF conference on computer vision and pattern recognition},
  pages={19679--19688},
  year={2022}
}

@article{xu2023murf,
  title={Murf: Mutually reinforcing multi-modal image registration and fusion},
  author={Xu, Han and Yuan, Jiteng and Ma, Jiayi},
  journal={IEEE transactions on pattern analysis and machine intelligence},
  volume={45},
  number={10},
  pages={12148--12166},
  year={2023},
  publisher={IEEE}
}

@inproceedings{ha2017mfnet,
  title={MFNet: Towards real-time semantic segmentation for autonomous vehicles with multi-spectral scenes},
  author={Ha, Qishen and Watanabe, Kohei and Karasawa, Takumi and Ushiku, Yoshitaka and Harada, Tatsuya},
  booktitle={2017 IEEE/RSJ International Conference on Intelligent Robots and Systems (IROS)},
  pages={5108--5115},
  year={2017},
  organization={IEEE}
}

@article{sun2022drone,
  title={Drone-based RGB-infrared cross-modality vehicle detection via uncertainty-aware learning},
  author={Sun, Yiming and Cao, Bing and Zhu, Pengfei and Hu, Qinghua},
  journal={IEEE TCSVT},
  volume={32},
  number={10},
  pages={6700--6713},
  year={2022},
  publisher={IEEE}
}

@inproceedings{xu2020aaai,
title={FusionDN: A Unified Densely Connected Network for Image Fusion},
author={Xu, Han and Ma, Jiayi and Le, Zhuliang and Jiang, Junjun and Guo, Xiaojie},
booktitle={proceedings of the Thirty-Fourth AAAI Conference on Artificial Intelligence},
year={2020}
}

@inproceedings{zhao2023cddfuse,
  title={Cddfuse: Correlation-driven dual-branch feature decomposition for multi-modality image fusion},
  author={Zhao, Zixiang and Bai, Haowen and Zhang, Jiangshe and Zhang, Yulun and Xu, Shuang and Lin, Zudi and Timofte, Radu and Van Gool, Luc},
  booktitle={Proceedings of IEEE Conference on Computer Vision and Pattern Recognition},
  pages={5906--5916},
  year={2023}
}

@INPROCEEDINGS{9577298,
  author={Zamir, Syed Waqas and Arora, Aditya and Khan, Salman and Hayat, Munawar and Khan, Fahad Shahbaz and Yang, Ming-Hsuan and Shao, Ling},
  booktitle={2021 IEEE/CVF Conference on Computer Vision and Pattern Recognition (CVPR)}, 
  title={Multi-Stage Progressive Image Restoration}, 
  year={2021},
  volume={},
  number={},
  pages={14816-14826},
  doi={10.1109/CVPR46437.2021.01458}}

@article{TANG2022SuperFusion,
      title={SuperFusion: A versatile image registration and fusion network with semantic awareness},
      author={Tang, Linfeng and Deng, Yuxin and Ma, Yong and Huang, Jun and Ma, Jiayi},
      journal={IEEE/CAA Journal of Automatica Sinica},
      volume={9},
      number={12},
      pages={2121--2137},
      year={2022},
      publisher={IEEE}
}

@inproceedings{huang2022reconet,
  title={Reconet: Recurrent correction network for fast and efficient multi-modality image fusion},
  author={Huang, Zhanbo and Liu, Jinyuan and Fan, Xin and Liu, Risheng and Zhong, Wei and Luo, Zhongxuan},
  booktitle={European conference on computer Vision},
  pages={539--555},
  year={2022},
  organization={Springer}
}

@inproceedings{bian2026fusionregister,
  title={FusionRegister: Every Infrared and Visible Image Fusion Deserves Registration},
  author={Bian, Congcong and Ma, Haolong and Li, Hui and Shen, Zhongwei and Luo, Xiaoqing and Song, Xiaoning and Wu, Xiao-jun},
  booktitle={Proceedings of the IEEE/CVF Conference on Computer Vision and Pattern Recognition},
  pages={41551--41561},
  year={2026}
}

@inproceedings{wu2025every,
  title={Every SAM drop counts: Embracing semantic priors for multi-modality image fusion and beyond},
  author={Wu, Guanyao and Liu, Haoyu and Fu, Hongming and Peng, Yichuan and Liu, Jinyuan and Fan, Xin and Liu, Risheng},
  booktitle={Proceedings of the Computer Vision and Pattern Recognition Conference},
  pages={17882--17891},
  year={2025}
}

@inproceedings{zhong2025amdanet,
  title={AMDANet: Attention-Driven Multi-Perspective Discrepancy Alignment for RGB-Infrared Image Fusion and Segmentation},
  author={Zhong, Haifeng and Tang, Fan and Chen, Zhuo and Chang, Hyung Jin and Gao, Yixing},
  booktitle={Proceedings of the IEEE/CVF International Conference on Computer Vision},
  pages={10645--10655},
  year={2025}
}

@inproceedings{wang2025source,
  title={The Source Image is the Best Attention for Infrared and Visible Image Fusion},
  author={Wang, Song and Han, Xie and Kuang, Liqun and Wang, Boying and Chen, Zhongyu and Qiao, Zherui and Yang, Fan and Liu, Xiaoxia and Zhang, Bingyu and Wang, Zhixun},
  booktitle={Proceedings of the IEEE/CVF International Conference on Computer Vision},
  pages={13513--13522},
  year={2025}
}

@inproceedings{dai2025multi,
  title={Multi-Modal Synergistic Implicit Image Enhancement for Efficient Optical Flow Estimation},
  author={Dai, Weichen and Wu, Hexing and Weng, Xiaoyang and Zheng, Yuxin and Ming, Yuhang and Kong, Wanzeng},
  booktitle={Proceedings of the Computer Vision and Pattern Recognition Conference},
  pages={2173--2182},
  year={2025}
}

@inproceedings{zhu2025wavemamba,
  title={WaveMamba: Wavelet-driven mamba fusion for RGB-infrared object detection},
  author={Zhu, Haodong and Dong, Wenhao and Yang, Linlin and Li, Hong and Yang, Yuguang and Ren, Yangyang and Zhu, Qingcheng and Feng, Zichao and Li, Changbai and Lin, Shaohui and others},
  booktitle={Proceedings of the IEEE/CVF International Conference on Computer Vision},
  pages={11219--11229},
  year={2025}
}

@inproceedings{hu2025balancing,
  title={Balancing task-invariant interaction and task-specific adaptation for unified image fusion},
  author={Hu, Xingyu and Jiang, Junjun and Wang, Chenyang and Jiang, Kui and Liu, Xianming and Ma, Jiayi},
  booktitle={Proceedings of the IEEE/CVF International Conference on Computer Vision},
  pages={11262--11272},
  year={2025}
}

@inproceedings{bai2025task,
  title={Task-driven image fusion with learnable fusion loss},
  author={Bai, Haowen and Zhang, Jiangshe and Zhao, Zixiang and Wu, Yichen and Deng, Lilun and Cui, Yukun and Feng, Tao and Xu, Shuang},
  booktitle={Proceedings of the Computer Vision and Pattern Recognition Conference},
  pages={7457--7468},
  year={2025}
}

@misc{yolo11_ultralytics,
author = {Jocher, Glenn and Qiu, Jing and Chaurasia, Ayush},
license = {AGPL-3.0},
month = jan,
title = {{Ultralytics YOLO}},
url = {https://github.com/ultralytics/ultralytics},
version = {8.0.0},
year = {2023}
}

@article{li2026uncertainty,
  title={Uncertainty-aware Spatial-Frequency Registration and Fusion for Infrared and Visible Images},
  author={Li, Xingyuan and Xu, Haoyuan and Zhu, Xingyue and Ma, Jun and Zou, Yang and Jiang, Zhiying and Liu, Jinyuan},
  journal={arXiv preprint arXiv:2605.13049},
  year={2026}
}

\end{document}